\documentclass[10pt,twocolumn,letterpaper]{article}

\usepackage{wacv}
\usepackage{times}
\usepackage{epsfig}
\usepackage{graphicx}
\usepackage{amsmath}
\usepackage{amssymb}

\usepackage[square, sort, comma, numbers]{natbib}
\usepackage{comment}
\usepackage{threeparttable}
\usepackage{bm}
\usepackage{multirow}
\usepackage{amsthm}
\usepackage{color}
\usepackage{subfigure} 
\usepackage[rightcaption]{sidecap}

\usepackage{algorithmic}
\usepackage[ruled,linesnumbered]{algorithm2e}
\usepackage{adjustbox}

\usepackage{booktabs}       
\usepackage{enumitem}

\usepackage{mydefs}

\def\wacvPaperID{3} 

\wacvfinalcopy 

\ifwacvfinal
\def\assignedStartPage{1} 
\fi

\ifwacvfinal
\usepackage[breaklinks=true,bookmarks=false]{hyperref}
\else
\usepackage[pagebackref=true,breaklinks=true,colorlinks,bookmarks=false]{hyperref}
\fi

\ifwacvfinal
\else
\fi

\begin{document}

\title{Zero-Shot Recognition via Optimal Transport}

\author{Wenlin Wang$^{1}$, \text{Hongteng Xu}$^{2}$, \text{Guoyin Wang}$^{3}$, \text{Wenqi Wang}$^{4}$, \text{Lawrence Carin}$^{1}$ \\
$^{1}$Duke University, $^{2}$Renmin University of China, $^{3}$Amazon Alexa AI, $^{4}$Facebook Inc\\
{\tt\small \{wlwang616, hongtengxu313\}@gmail.com },
{\tt\small guoyiwan@amazon.com },
{\tt\small wenqiwang@fb.com },
{\tt\small lcarin@duke.edu} 
}

\maketitle

\begin{abstract}
We propose an optimal transport (OT) framework for generalized zero-shot learning (GZSL), seeking to distinguish samples for both seen and unseen classes, with the assist of auxiliary attributes. 
The discrepancy between features and attributes is minimized by solving an optimal transport problem. 
{Specifically, we build a conditional generative model to generate features from seen-class attributes, and establish an optimal transport between the distribution of the generated features and that of the real features.} 
The generative model and the optimal transport are optimized iteratively with an attribute-based regularizer, that further enhances the discriminative power of the generated features. 
A classifier is learned based on the features generated for both the seen and unseen classes. 
In addition to generalized zero-shot learning, our framework is also applicable to standard and transductive ZSL problems.
Experiments show that our optimal transport-based method outperforms state-of-the-art methods on several benchmark datasets.
\end{abstract}

\section{Introduction}
When there is access to abundant labeled examples for image-based data, modern machine learning and deep learning algorithms have demonstrated the ability to learn reliable classifiers ~\cite{he2016deep,szegedy2015going,simonyan2014very}. 
Unfortunately, their ability to generalize to unseen classes typically remains poor. 
This limitation has motivated significant recent interest in zero-shot learning (ZSL)~\cite{socher2013zero,lampert2014attribute,vinyals2016matching,ravi2016optimization}. 
By leveraging auxiliary information that may be available for the seen/unseen classes, $e.g.$, attribute vectors and/or textural descriptions of classes~\cite{mikolov2013distributed}, ZSL aims to learn new concepts with minor or no supervision ($i.e.$, distinguish data for classes unseen in the training phase). 

Although many ZSL methods have been proposed, they often suffer from inherent limitations. 
A typical problem is ``\textit{domain shift}.''
Many ZSL methods try to establish a mapping between the feature space and the class/attribute space, and predict the unseen classes by finding their closest attribute vectors~\cite{akata2013label,lampert2014attribute}. 
However, seen classes and unseen ones often suffer from clear domain differences in high-dimensional space.
Accordingly, the mapping learned based on the seen classes may be inapplicable to the unseen classes. 
Another problem is ``\textit{high-bias}.''
Most methods highly bias towards predicting the seen classes~\cite{verma2018generalized,xian2018feature}, because the training data are purely from the seen classes. 
Such a phenomenon is important in generalized zero-shot learning (GZSL), where the proposed classifier needs to distinguish samples for both seen and unseen classes. 

\begin{figure*}[t]
    \centering
    \includegraphics[width=0.85\linewidth]{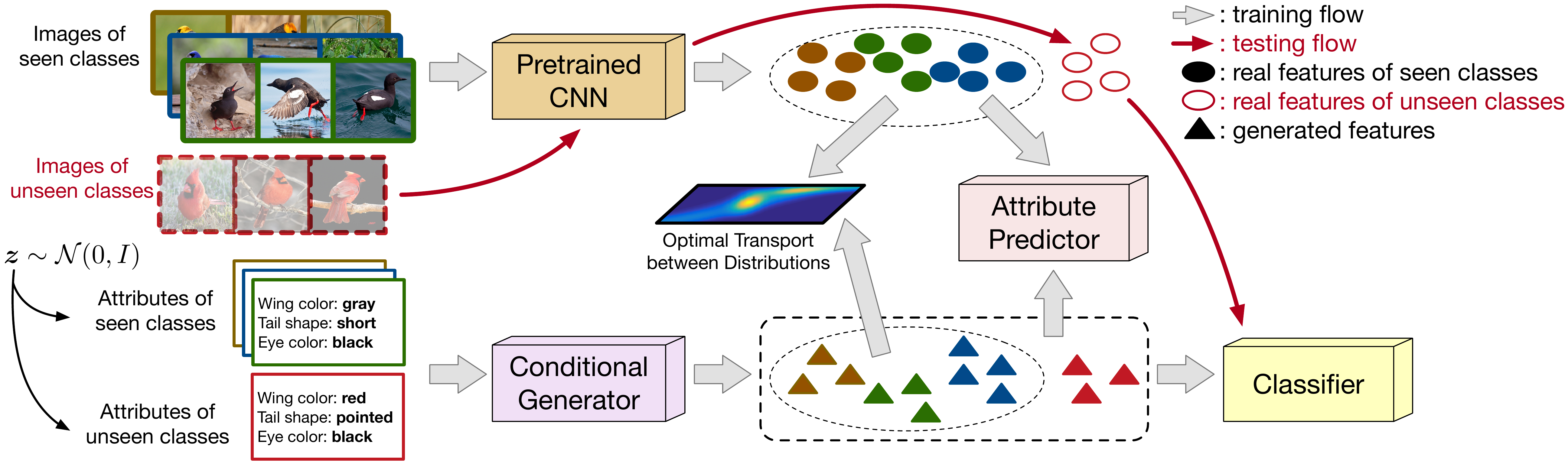}
    \caption{The diagram of proposed method. 
    In the training phase, a generator is learned to synthesize features from attributes, whose distribution approaches to that of real features obtained from a pretrained CNN. 
    After the generator is learned, we further train a classifier based on the generated features for both seen and unseen classes. 
    In the testing phase, the ``pretrained CNN + classifier'' achieves zero-shot learning, whose pipeline is shown as red arrows.} 
    \label{fig:main_arch}
\end{figure*}

To overcome the above problems, we propose an optimal-transport-based zero-short learning method. 
The proposed method is motivated by recent {\em synthesis} of exemplars for ZSL~\cite{verma2018generalized,mishra2017generative, xian2018feature}, extended by building a more powerful generative model via optimal transport (OT)~\cite{villani2008optimal}. 
As shown in Figure~\ref{fig:main_arch}, our method learns a conditional generative model to construct features from attributes, and minimizes the optimal transport distance (also called Wasserstein discrepancy) between the generated features and the real ones. 
The generative model helps to synthesize samples for unseen classes from given attributes. 
We apply iterative optimization to jointly learn the generative model and the corresponding optimal transport, and further enhance the discriminative power of the generated data via learning an associated attribute predictor as a regularization. 
Finally, an additional classifier is trained to distinguish generated data from both seen and unseen classes.

Distinct from existing methods, which minimize the discrepancy between features and attributes on each individual data, our method considers the optimal transport between the {\em distribution} of features and that of attributes.   
{Matching distributions~\cite{tsai2017learning, xian2018feature} is found to be more robust than finding correspondences between individual samples, with this beneficial for suppressing the risk of domain shift caused by mismatch.} 
Additionally, the generative model can synthesize instances for unseen classes, with the synthesized data available for training the classifier.  Therefore, the \textit{high-bias} problem can be suppressed effectively. 
Moreover, our method is the first framework conducting ZSL under the primal form of optimal transport (OT), and it provides universal solutions to ZSL problems. 
It is applicable not only to the generalized ZSL~\cite{xian2017zero} problem, but also to the standard ZSL~\cite{lampert2014attribute} and transductive ZSL~\cite{rohrbach2013transfer}. 
We analyze the assumptions of our method in depth and investigate its robustness to hyper-parameters. 
Experimental results demonstrate that the proposed approach outperforms state-of-the-art methods on several image-based benchmark datasets. 

\section{Proposed Framework}
Suppose there are $S$ seen classes and $U$ unseen classes. 
For the $i$-th class, $i\in\{1,...,S+U\}$, we are given a $d$-dimensional attribute vector~\cite{socher2013zero}, denoted $\av_i\in\mathbb{R}^d$. 
All the attribute vectors formulate an attribute matrix $\Av=[\av_i]\in\mathbb{R}^{d\times (S+U)}$. 
Data from the same class share the same attributes and different classes always have different attributes. 
The seen classes contain $N$ labeled samples $\mathcal{D}_s = \{ (\xv_n, \av_{n})\}_{n=1}^N$, where $\xv_n\in \mathbb{R}^D$ and $\av_n \in \{\av_1,...,\av_S\}$ represent the $n$-th sample and its attribute/label, respectively.
For the $U$ unseen classes, the unlabeled data are denoted $\mathcal{D}_u =\{\xv_{n+N}\}_{n=1}^{N'}$, whose correspondences with $\{\av_{S+1},...\av_{S+U}\}$ are unknown. In the work considered here, $\xv_n$ corresponds to features extracted from image $n$, and such features will be manifested by a deep neural network operating on the image.

We consider three settings for zero-shot learning:  standard, generalized and transductive. 
In the standard setting~\cite{lampert2014attribute}, we only have access to $\mathcal{D}_s$ and the attribute matrix $\Av$ in the training phase, and focus on classification of the samples in $\mathcal{D}_u$ in the testing phase.
Generalized ZSL~\cite{xian2017zero} requires the final classifier to categorize the samples in the whole $U+S$ classes. 
Similar to the standard setting, the transductive ZSL~\cite{kodirov2015unsupervised} also aims to train a classifier for the $U$ unseen classes, but it uses both the labeled data $\mathcal{D}_s$ and the unlabeled data $\mathcal{D}_u$ in the training phase.

We solve these three ZSL tasks within a unified framework. 
The key of the proposed approach is to learn a conditional generative model ($i.e.$, the ``\textit{generator}'' in Figure~\ref{fig:main_arch}), capable of generating representative features for both seen and unseen classes, conditioned on their attributes. 
Those generated features are used to train a classifier.\footnote{In standard and transductive ZSL problems, the classifier is just for the $U$ unseen classes. In the generalized ZSL problem, the classifier is for all $U+S$ classes.}

The quality of the generated features has a significant influence on the target classifier, which is purely decided by the proposed generator. 
According to the nature of ZSL problems, it is necessary for an ideal generator to have the following properties:

\vspace{5pt}
\emph{1. For seen classes, the distribution of their generated features should be close to that of real features.} 
\vspace{5pt}

\emph{2. For both seen and unseen classes, the generated features should have discriminative power, $e.g.$, the features from the same attribute have a clustering structure.}
\vspace{5pt}

The first property means that the generator establishes a reliable mapping from attributes to features. 
The second property ensures that the features synthesized by the generator are informative to train a classifier. 
The generator with these two properties can synthesize realistic features and be extended to those unseen classes. 
Since the generator can synthesize a large quantity of features for unseen classes when training the classifier, the bias between seen and unseen classes can be suppressed.
Guided by these two properties, we design an optimal transport-based method to learn the generator.

\section{Conditional Feature Generator}
\subsection{Optimal transport-based loss}
The conditional feature generator is denoted $g:~\mathcal{A}\mapsto \mathcal{X}$, with $\mathcal{A}$ and $\mathcal{X}$ the attribute and feature space, respectively. 
Given an attribute vector $\av\in\mathcal{A}$, we generate a set of synthetic features via
\begin{eqnarray}\label{eq:generator}
\begin{aligned}
\hat{\xv}=g([\av; \zv]),~\zv\sim \mathcal{N}(\bm{0}, \bm{I}_d),
\end{aligned}
\end{eqnarray}
where $\zv\in \mathbb{R}^d$ is a random variable drawn from a normal distribution, and $[\av; \zv]$ denotes the concatenation of vectors $\av$ and $\zv$. We choose to make the dimension of $\zv$ the same as that of $\av$; while not required, it was found to work well in practice.

For the seen classes, we have $N$ real features $\{\bm{x}_n\}_{n=1}^{N}$,\footnote{In this work, we focus on the zero-shot learning problem in image classification. 
Taking images as input, a pre-trained convolutional neural network outputs the real features.}
and obtain $M$ generated features $\{\hat{\xv}_m\}_{m=1}^{M}$ from the attributes of the seen classes via (\ref{eq:generator}). 
Instead of finding correspondences between the features generated from the attribute space and those in the feature space, we aim to match the empirical {\em distribution} of the generated features to that of real features (as mentioned in Property 1). 
We utilize distribution matching because point-wise matching in high-dimensional space is sensitive to fluctuations of samples and to outliers, yielding solutions that often fall into bad local optima with incorrect correspondences. 
Distribution matching, on the contrary, considers more variability of the data and matches the data globally. 
In practice, distribution matching has proven to be effective for transferring knowledge from one modality to another~\cite{long2015learning,tsai2017learning,hubert2016learning}.

To measure the distance between distributions, the optimal transport (OT) distance~\cite{villani2008optimal} is a natural choice. 
Mathematically, the optimal transport distance between two probability measures $\muv$ and $\vv$ is defined as~\cite{peyre2017computational}:
\begin{eqnarray}\label{eq:ot}
\begin{aligned}
    d_c(\muv, \vv) = \sideset{}{_{\pi\in \Pi (\muv, \vv)}}\inf \mathbb{E}_{(\xv, \xv')\sim \pi} [c(\xv, \xv')]
\end{aligned}
\end{eqnarray}
where $\Pi(\muv, \vv)$ is the set of all joint distributions with $\muv$ and $\vv$ as marginals, and $c(\xv, \xv')$ is the cost for moving from $\xv$ to $\xv'$. 
When the cost is defined as the Euclidean distance, $i.e.$, $\Vert \xv-\xv' \Vert_2$, (\ref{eq:ot}) corresponds to the Wasserstein distance. 
When the cost is not a metric, which is common in practice, (\ref{eq:ot}) is also called Wasserstein discrepancy. 

In our case, both the {\em empirical} distribution of real features and that of generated features are represented as uniformly distributed, denoted as $\muv=[\frac{1}{N}] \in \mathbb{R}^N$ and $\vv=[\frac{1}{M}]\in \mathbb{R}^M$, and (\ref{eq:ot}) can be rewritten as
\begin{eqnarray}\label{eq:ot2}
\begin{aligned}
\mathcal{L}_{OT}(\mathcal{D}_s, \Av; g) &= \sideset{}{_{\Tv\in \Pi (\muv, \vv)}}\min \sideset{}{_{n}}\sum\sideset{}{_{m}}\sum T_{nm} C_{nm}\\
&=\sideset{}{_{\Tv\in \Pi (\muv, \vv)}}\min \mbox{tr}(\Tv^{\top} \Cv).
\end{aligned}
\end{eqnarray}
Here, $\Pi(\muv, \vv)=\{\Tv|\Tv \bm{1}_{M} = \muv, \Tv^{\top}\bm{1}_N = \vv\}$. 
The cost $C_{nm}=C(\xv_n, \hat{\xv}_m)$ is the distance between the $n$-th real feature and the $m$-th generated feature, which can be designed with high flexibility. 
Since the usage of Euclidean distance in a high dimensional embedding space will lead to a severe \textit{hubness} problem~\cite{dinu2014improving}, we define the cost as the cosine distance $C_{nm} = 1 - \frac{\xv_n^{\top} \hat{\xv}_m}{\Vert\xv_n\Vert_2 \Vert\hat{\xv}_m\Vert_2}$. 
$\Tv=[T_{nm}]\in \mathbb{R}^{N\times M}$ is the proposed optimal transport matrix. 
The element $T_{nm}\geq 0$ denotes the probability that the real feature $\xv_n$ matches with the generated feature $\hat{\xv}_m$. 

The optimal transport distance supplies an unsupervised alignment between the empirical distribution of real features and that of synthetic features generated from attributes. By minimizing this distance, we can learn a generator to synthesize reasonable features consistent with the real data distribution. 
Compared with other distances, $e.g.$, KL-divergence~\cite{edwards2016towards, wang2017zero} and maximum mean discrepancy (MMD)~\cite{tsai2017learning}, optimal transport distance has some advantages. 
Specifically, KL-divergence is asymmetric and may suffer from a ``vanishing gradient'' problem when the supports of two distributions are non-overlapped~\cite{chen2018symmetric} (which is common when our generator is initialized randomly). 
MMD just considers the similarity of the first order statistic, whose constraints are too loose for distribution matching. 
The proposed optimal transport distance, however, is applicable to non-overlapped distributions, which suppresses the ``vanishing gradient'' problem when learning our generator. 
Additionally, the optimal transport distance imposes more constraints on the generator than MMD does, which accordingly has lower risk of over-fitting.

\subsection{Attribute-based regularizer}~\label{sec:attribute_regularizer}
By minimizing the optimal transport distance, $i.e.$,  $\min_{g}\mathcal{L}_{OT}(\mathcal{D}_s, \Av; g)$, we ensure that the generator can mimic the distribution of features based on the corresponding attributes. 
However, there is no guarantee that the generated features are discriminative enough to train a good classifier. 
To reduce this potential risk, we propose an attribute-based regularizer. 
In particular, given a feature vector, either the real $\xv\in\mathcal{D}_s$ or the synthetic $\hat{\xv}$ from our generator, the proposed regularizer aims to maximize its conditional likelihood given the corresponding attribute. 
Taking advantage of neighborhood component analysis (NCA)~\cite{goldberger2005neighbourhood}, we define the conditional likelihood of a feature $\xv$ (or $\hat{\xv}$) given an attribute $\av$ as
\begin{eqnarray}\label{eq:prob}
\begin{aligned}
    p(\xv|\av) = \frac{ \exp(-\gamma^2 d(f({\xv}), \av) )}{\sum_{i=1}^{S+U}\exp(-\gamma^2 d(f({\xv}), \av_{i}) )},
\end{aligned}
\end{eqnarray}
where $f: \mathcal{X}\mapsto\mathcal{A}$ is a predictor estimating the attributes of generated features, and $\gamma^2$ is a hyper-parameter tuning the strength of the discriminator power. 
The discrepancy between the proposed attribute and the predicted one is defined by the cosine similarity $d(f({\xv}), \av) = 1 - \frac{f({\xv})^{\top} \av}{\Vert f({\xv})\Vert_2 \Vert \av\Vert_2}$. 

Accordingly, we calculate negative log-likelihood of both the labeled features from the seen classes and those generated from both seen and unseen classes, and impose it on our generator as a regularizer: 
\begin{eqnarray}\label{eq:lle}
\begin{aligned}
    &\mathcal{L}_P(\mathcal{D}_s,\Av;g,f) \\
    &=  -\mathbb{E}_{\xv,\av\sim \mathcal{D}_s}[\log p(\xv|\av) ] -\mathbb{E}_{\hat{\xv}}[\log p(\hat{\xv}|\av) ] \\
    &= -\mathbb{E}_{\xv,\av\sim \mathcal{D}_s}[\log p(\xv|\av) ]
    -\mathbb{E}_{\zv}[\log p(g([\av;\zv])|\av)].
\end{aligned}
\end{eqnarray}
By minimizing (\ref{eq:lle}), the generator is endowed discriminative power for the generated features. 
Further, this regularizer provides guidance for the learning of the generative model, navigating the generator to construct high-quality instances yielding more rational transport.

Combining the optimal-transport-based loss with the attribute-based regularizer, we optimize the proposed feature generator $g$ associated with the attribute predictor $f$ via
\begin{eqnarray}\label{eq:obj}
\begin{aligned}
    \sideset{}{_{g,f}}\min \mathcal{L}_{OT}(\mathcal{D}_s,\Av;g) + \beta \mathcal{L}_P(\mathcal{D}_s,\Av;g,f)
\end{aligned}
\end{eqnarray}
where $\beta$ is a hyper-parameter weighting the optimal transport loss and the attribute predictor. 

\section{Learning Algorithm}
The optimization problem~(\ref{eq:obj}) can be solved by an iterative optimization strategy. 
Specifically, on each iteration we first calculate the optimal transport distance given the cost derived from the current generator $g$, and then optimize $g$ and $f$ based on estimated optimal transport matrix $\Tv$. 

\subsection{Calculating optimal transport distance}
Given the generator $g$ obtained in the previous iteration, we generate $M$ features $\{\hat{\xv}_m\}_{m=1}^M$ and calculate the cost matrix $\Cv$ accordingly. 
The optimal transport distance $\mathcal{L}_{OT}$ can then be calculated by solving (\ref{eq:ot}). 
Instead of applying linear programming (LP) directly to solve (\ref{eq:ot}), whose complexity is $\mathcal{O}(N^3\log N)$, in this work we calculate the optimal-transport distance using the Inexact Proximal point method for Optimal
Transport (IPOT)~\cite{xie2018fast}. 
This method finds the optimal transport $\Tv$ iteratively, and in each iteration it imposes a Bregman divergence term $D_B$ to (\ref{eq:ot}).
Accordingly, the optimization problem becomes
\begin{eqnarray}\label{eq:ipot}
\begin{aligned}
    \Tv^{(t+1)} =  \sideset{}{_{\Tv\in \Pi (\muv, \vv)}}\min \mbox{tr}(\Tv^{\top}\Cv) + \lambda D_B (\Tv, \Tv^{(t)}), 
\end{aligned}
\end{eqnarray}
where $D_B(\Tv, \Tv^{(t)}) = \sum_{n,m}T_{nm} \log\frac{T_{nm}}{T^{(t)}_{nm}}$ calculates the Kullback-Leibler (KL) divergence between the proposed $\Tv$ and its estimation in the $t$-th iteration; $\lambda$ controls the significance of the regularizer. 

Equation (\ref{eq:ipot}) can be rewritten as
\begin{eqnarray}\label{eq:sinkhorn}
\begin{aligned}
\sideset{}{_{\Tv\in \Pi (\muv, \vv)}}\min  \mbox{tr}(\Tv^{\top}(\Cv-\log\Tv^{(t)})) + \lambda H(\Tv), 
\end{aligned}
\end{eqnarray}
where $H(\Tv) = \sum_{n,m} T_{nm}\log T_{nm}$ is the entropy term of $\Tv$. 
Equation (\ref{eq:sinkhorn}) is of the same form as the Sinkhorn distance, introduced in~\cite{cuturi2013sinkhorn}, which can be solved with near linear complexity. 
The solution to (\ref{eq:sinkhorn}) can be obtained efficiently via the Sinkhorn-Knopp algorithm~\cite{sinkhorn1967concerning}.

The IPOT~\cite{xie2018fast} algorithm is employed in our framework because it has several advantages compared with traditional linear programming-based algorithms~\cite{pele2009fast} and Sinkhorn iteration~\cite{cuturi2013sinkhorn}. 
First, the per-iteration computational complexity of IPOT is at most comparable to that of Sinkhorn iteration, and is much lower than that of linear programming. 
Secondly, although both Sinkhorn iteration and IPOT have near-linear convergence, IPOT requires much fewer inner iterations, because it only needs to find inexact proximity in each step. 
Thirdly, the Sinkhorn iteration algorithm is sensitive to the choice of the regularizer's weight, while the IPOT method is robust to the change of the weight in a wide range.
More detailed analysis of the IPOT method can be found in our supplementary file and related references~\cite{pele2009fast, cuturi2013sinkhorn, xie2018fast}.

\subsection{Modifications for ZSL problems}~\label{sec:stochastic_transition}
Note that (\ref{eq:ot2}) provides an unsupervised solution to align the generated data distribution and the empirical data distribution. 
For the seen classes, however, the correspondence between these data can also be derived by comparing the attribute vectors of the real data with the generator's inputs. 
To leverage the correspondence provided by attributes, we propose a \textit{stochastic transition} method when learning the optimal transport. 
In particular, when learning the parameters of our generative model, we have a probability $p$ to learn the optimal transport matrix $\Tv^*$ defined in (\ref{eq:ipot}), and $1-p$ to directly assign the transition matrix $\widetilde{\Tv}$ with supervised signal. 
Given the real feature $\xv_{n}$ associated with the attribute $\av_{n}$ and the synthetic feature $\xv_m =g([\av_m; \zv_m])$,\footnote{Here $\av_m\in \{\av_1,...,\av_S\}$, which is randomly selected from the attributes of seen classes.} the element of $\widetilde{\Tv}$ is defined as 
\begin{eqnarray}
\label{eq:tildeT}
\begin{aligned}
\widetilde{T}_{mn}= 
\begin{cases}
    \frac{1}{\#\{\av_{n}\} \times \#\{\av_{m}\}},& \text{if } \av_{n}=\av_m\\
    0,              & \text{otherwise},
\end{cases}
\end{aligned}
\end{eqnarray}
where $\#\{\av\}$ counts the number of an attribute appearances. 
$\widetilde{\Tv}$ is a valid transport matrix in $\Pi(\muv, \vv)$, and provides guidance to align the two distributions in the image feature space.  Practically, this method supplies minor improvements for classification, but speeds learning substantially. 

\subsection{Updating the generative model}
Given the optimal transport matrix $\Tv^*$, we further update the generator $g$ and predictor $f$ by
\begin{eqnarray}\label{eq:subproblem}
\begin{aligned}
\sideset{}{_{g, f}}\min \mbox{tr}(\Tv^{*\top}\Cv_g) + \beta \mathcal{L}_P(\mathcal{D}_s, \Av; g, f), 
\end{aligned}
\end{eqnarray}
where $\Cv_g$ is the cost matrix parameterized by the generator.
This problem can be solved efficiently via stochastic gradient descent. 
We apply Adam~\cite{kingma2014adam} to update $g$ and $f$. 
The whole learning process is summarized in Algorithm~\ref{alg2}.

\begin{algorithm}[t]
	\caption{Iterative Optimization}
	\label{alg2}
	\begin{algorithmic}[1]
		\STATE \textbf{Input:} $\mathcal{D}_s = \{(\xv_n, \av_{n})\}_{n=1}^N$, attribute matrix $\Av$, $p=0.9$, $\gamma^2 =0.5$, $b=128$
		\STATE \textbf{Output:} $g(\cdot)$
		\WHILE{not converge}
		\STATE Sample $B_{real}=\{(\xv_i, \av_i)\}_{i=1}^b$ from $\mathcal{D}_s$.
		\STATE For seen classes, sample $B_{s}=\{(\hat{\xv}_i, \av_i)\}_{i=1}^b$, $\av_i\in\{\av_1,..,\av_S\}$, via~(\ref{eq:generator}).
		\STATE For unseen classes, sample $B_{u}=\{(\hat{\xv}_i, \av_i)\}_{i=1}^b$, $\av_i\in \{\av_{S+1},...,\av_{S+U}\}$, via~(\ref{eq:generator}).
		\STATE Sample $z\sim\mbox{Uniform}[0,1]$.
		\IF{$z\le p$}
		\STATE Based on $B_{real}$ and $B_s$, update $\Tv \leftarrow \Tv^{*}$ via (\ref{eq:ipot})
		\ELSE
		\STATE $\Tv \leftarrow \tilde{\Tv}$ via (\ref{eq:tildeT}).
		\ENDIF
		\STATE Based on $B_{s}$ and $B_{u}$, solve (\ref{eq:subproblem}) via SGD and update $\{g, f\}$ accordingly.
		\ENDWHILE
	\end{algorithmic}
\end{algorithm}

\subsection{Classifier}
Given the generative model learned in the previous section, we are able to generate representative features of both seen and unseen classes by (\ref{eq:generator}). 
These generated samples, together with the data from the seen classes, can be used to train a classifier, $e.g.$, the simple linear softmax function used in the following experiments. 
Since the classifier leverages labeled examples from both seen and (synthesized) unseen classes, the corresponding ZSL result is inherently robust against bias towards seen classes. 

\section{Related Work}
\textbf{Wasserstein GAN (WGAN)}~\cite{arjovsky2017wasserstein}. 
The proposed optimal-transport-based method is akin to a WGAN model~\cite{xian2018feature}, with some key differences. 
WGAN~\cite{arjovsky2017wasserstein} and its variants~\cite{gulrajani2017improved} use Kantorovich-Rubinstein duality to calculate the Wasserstein distance. 
A constraint for the dual form is that the discriminator must be a 1-Lipschitz function, which may be violated in practice. 
In our model, we solve the optimal transport problem in its prime form directly. 
As a result, our method does not play the max-min game like GAN-related work does, and does not need to train an additional discriminator to distinguish real and synthetic features.
Additionally, Wasserstain distance is a special case for the optimal transport distance~\cite{villani2008optimal}. 
Our method can be easily adapted to other metrics. 

\textbf{Denoising Auto-Encoder (DAE)}~\cite{vincent2010stacked}. 
The proposed generator together with an attribute-based regularizer is similar to a DAE~\cite{vincent2010stacked}, by reconstructing the corresponding attribute vector from the noised input attribute. 
However, our model is specialized for classification tasks. 
In particular, we introduce a neighborhood component analysis (NCA)~\cite{goldberger2005neighbourhood} loss as the regularizer. 
Essentially, this regularizer implies that the features should have clustering structure defined by the corresponding attributes. 
Therefore, the features generated by our model are more discriminative, suitable for training the following classifier.

\textbf{OT-GAN}~\cite{salimans2018improving}.
Our model is similar to OT-GAN~\cite{salimans2018improving} in spirit, learning optimal transport in the primal space, however, different in both modeling and algorithm. OT-GAN~\cite{salimans2018improving} is designed for image generation that cannot be directly extended for ZSL problems, while our OT framework serves for ZSL tasks, with numbers of specialized modules, \textit{e.g.}, the stochastic transition method (Sec~\ref{sec:stochastic_transition}) and the attribute regularizer (Sec~\ref{sec:attribute_regularizer}). These modules are essential for ZSL.  Additionally, OT-GAN~\cite{salimans2018improving} applies entropic regularizer and learns optimal transport via the Sinkhorn algorithm~\cite{cuturi2013sinkhorn}, while our work uses KL-divergence-based regularizer and the proximal gradient method (\textit{i.e.}, IPOT~\cite{xie2018fast}), which is more robust to hyperparameters and with more stable convergence. 

\textbf{Zero-shot learning methods}
From the viewpoint of methodology, ZSL methods can be roughly categorized into five types:
($i$) Learning a mapping from the feature space to the attribute space, and predicting the class of an unseen class test instance by finding its closest class-attribute vector~\cite{socher2013zero, lampert2014attribute, akata2015evaluation, chen2018zero}; 
($ii$) Learning a ``reverse'' projection from the attribute space to the feature space~\cite{zhang2016learning, li2018discriminative}, which improves the robustness of nearest-neighbor search; 
($iii$) Representing the classifier for each unseen class as a weighted combination of those for the seen classes, with the combination weights defined by a similarity score of unseen and seen class~\cite{zhang2015zero, changpinyo2016synthesized}; 
($iv$) Leveraging a probability distribution for each seen class and extrapolating to those unseen classes~\cite{verma2017simple,wang2017zero,mishra2017generative}. 
($v$) Building on a knowledge graph to predict the image categories~\cite{wang2018zero, lee2018multi}. 
All these types of methods have been widely used in the standard and transductive ZSL problems~\cite{fu2015transductive,li2015semi,rohrbach2013transfer}.
Recently, generalized zero-shot learning (GZSL)~\cite{chao2016empirical,xian2017zero} has been demonstrated to be a more challenging task. 
Among existing ZSL methods, generative models~\cite{kingma2013auto, goodfellow2014generative} have achieved significant success, including VAEs~\cite{verma2017simple,mishra2017generative} and GANs~\cite{xian2018feature}.
Our method is derived from generative models and applicable to various ZSL problems. Different from prior work, our framework seeks to generate data by minimizing the Wasserstein distance in the primal space. 

\textbf{Optimal transport and Wasserstein distance}
Optimal transport and Wasserstein learning have proven useful in distribution estimation~\cite{boissard2015distribution}, alignment~\cite{zemel2017fr} and clustering~\cite{agueh2011barycenters,ye2017fast,cuturi2014fast}, avoiding over-smoothed intermediate interpolation results.
The lower bound of Wasserstein distance has been used as a loss function when learning generative models~\cite{courty2017learning,arjovsky2017wasserstein}.
The main bottleneck of the application of optimal transport is its high computational complexity.
This problem has been greatly eased since the Sinkhorn iterative algorithm was proposed in~\cite{cuturi2013sinkhorn}, which applies iterative Bregman projection~\cite{benamou2015iterative} to approximate Wasserstein distance, and achieves a near-linear time complexity~\cite{altschuler2017near}. 
Many more complicated models have been proposed based on Sinkhorn iteration~\cite{genevay2017sinkhorn,schmitz2017wasserstein} and its variants, $e.g.$, Greenkhorn iteration~\cite{altschuler2017near} and IPOT~\cite{xie2018fast}. 
Motivated by these prior work, our framework solves the optimal transport problem with IPOT algorithm~\cite{xie2018fast} and demonstrates its superiority on real datasets in Sec~\ref{Sec: IPOTvsSinkhorn}.

\section{Experiments}
To evaluate the effectiveness of our method (denoted as \textbf{OT-ZSL} for optimal transport-based zero-shot learning), we apply it to generalized ZSL (GZSL), standard ZSL (SZSL) and transductive ZSL (TZSL), and compare it with state-of-the-art methods. 
Additionally, to investigate the functionality of each module in our method, we also consider a variant of our method, which is trained without the attribute-based regularizer, and denoted \textbf{OT-ZSL (w/o $\fv$)}.

\subsection{Image datasets and implementation details}
We report results on the following datasets, with associated detailed information found in Table~\ref{tab:dataset}.
\begin{itemize}[noitemsep,topsep=0pt,parsep=0pt,partopsep=0pt]
\item \textbf{Animals with Attributes (AwA)}~\cite{lampert2014attribute} This is a coarse-grained dataset containing 30,475 images, with 50 classes and 85 attributes. 
A standard split of 40 seen classes and 10 unseen classes are provided. 
Recently, an alternative data split is also available~\cite{xian2017zero}. 
We refer to the data with the original split as AwA1~\cite{lampert2014attribute}, and with the new split as AwA2~\cite{xian2017zero}.
\item \textbf{Caltech-UCSD-Birds-200 (CUB)}~\cite{wah2011caltech} This dataset consists of 11,788 fine-grained bird images, with 200 classes in total. 
A split of 150 unseen and 50 seen classes are provided. 
Following~\cite{verma2018generalized}, class attributes are obtained by averaging all the image level attributes.
\item \textbf{SUN Scene Recognition}~\cite{xiao2010sun} SUN contains 14,340 images from 717 scenes annotated with 102 attributes. 
We follow the most widely employed setting, with 645 seen classes and 72 unseen classes. 
\item \textbf{ImageNet}~\cite{russakovsky2015imagenet} Following~\cite{fu2016semi}, 1000 classes from ILSVRC-20112~\cite{russakovsky2015imagenet} are used as the seen classes, while 360 non-overlapped classes of ILSVRC-2010~\cite{deng2009imagenet} are used as unseen classes.
\end{itemize}

\begin{table}[t]
	\centering
	\caption{\small 
		Basic information of the considered datasets.
	}\label{tab:dataset}
	\begin{small}
	\begin{threeparttable}[c]
	\begin{tabular}{
	    @{\hspace{4pt}}c@{\hspace{4pt}}|
	    @{\hspace{4pt}}c@{\hspace{4pt}}
        @{\hspace{4pt}}c@{\hspace{4pt}}
        @{\hspace{4pt}}c@{\hspace{4pt}}
        @{\hspace{4pt}}c@{\hspace{4pt}}
        @{\hspace{4pt}}c@{\hspace{4pt}}
	}
	\hline\hline
	   Dataset &$\av$ &$d$ &$\#$Image & $\#S$ train/val. & $\#U$\\
		\hline
		AwA1& Attribute & 85 & 30,475 & $27/13$ &$10$ \\ 
		AwA2& Attribute & 85 & 37,322 & $27/13$ &$10$ \\
		CUB & Attribute & 312& 11,788 & $100/50$ & $50$\\
		SUN & Attribute & 102& 14,340 & $580/65$ &$72$\\
		ImageNet & Word2Vec & 1000 & 254,000 & $800/200$ & $360$ \\ 
	\hline\hline
	\end{tabular}
    \end{threeparttable}
    \end{small}
    \vspace{-1em}
\end{table}

\textbf{Features} 
For AwA1, AwA2, CUB and SUN, we extract 2048-dimensional features from a pre-trained 101-layered ResNet~\cite{he2016deep}. 
Their class attributes are the corresponding attribute vectors in these four datasets. 
For ImageNet, to make a fair comparison to previous work, we maintain the usage of GoogleNet~\cite{szegedy2015going}, which yields a 1024-dimensional extracted feature.
Its class attributes are the semantic word vector obtained from word2vec embeddings~\cite{mikolov2013distributed}.

\textbf{Implementation} 
For the reported experiments, we use the proposed train/test split~\cite{xian2017zero} for each dataset for GZSL, and consider both the standard and the proposed train/test split~\cite{xian2017zero} for SZSL. 
To make a fair comparison to prior work, only the standard split is evaluated for TZSL. 
For the network architecture, MLP with ReLU activation is used for both the feature generator and the attribute predictor. 
In all experiments, we use a single hidden layer with 4,096 hidden units. 
The noise $\zv$ is sampled from normal distribution $\mathcal{N}(\bm{0}, \bm{I}_d)$. 
Finally, for each seen/unseen class, we draw 500 synthetic features from our generator, and train a linear softmax classifier. 
Following~\cite{xian2018feature}, we divided the training data into training set and validation set, as shown in Table~\ref{tab:dataset}. 
We set hyper-parameters empirically based on the performance over the validation set. 
In all experiments below, we set $p=0.9$, $\lambda=0.5$, $\gamma^2=0.5$, $\beta=0.05$, and apply Adam~\cite{kingma2014adam}, with batch in size of 128 and learning rate in $0.001$, to train our model.

\begin{table*}[t]
	\centering
	\caption{\small 
	   Comparisons for various methods in the generalized ZSL problem, on the split provided by~\cite{xian2017zero}.
	}\label{res:GZSL}
	\begin{small}
	\begin{tabular}{l|ccc|ccc|ccc|ccc}
	\hline\hline
	     \multirow{2}{*}{Methods} & \multicolumn{3}{c|}{AwA1}& \multicolumn{3}{c|}{AwA2}& \multicolumn{3}{c|}{CUB} & 
	     \multicolumn{3}{c}{SUN} \\
		 &$A_u$  &$A_s$  &$H$  
		 &$A_u$  &$A_s$  &$H$ 
		 &$A_u$  &$A_s$  &$H$
		 &$A_u$  &$A_s$  &$H$\\
		\hline
		SJE~\cite{akata2015evaluation}  & 11.3& 74.6& 19.6&    8.0& 73.9& 14.4&    23.5& 59.2& 33.6&    14.7& 30.5& 19.8 \\
		LATEM~\cite{xian2016latent}     & 7.3& 71.7& 13.3&    11.5& 77.3& 20.0&    15.2& 57.3& 24.0&    14.7& 28.8& 19.5 \\
		SSE~\cite{zhang2016learning}    & 7.0& 80.5& 12.9   &8.1& 82.5& 14.8&    8.5& 46.9& 14.4&    2.1& 36.4& 4.0 \\
		ESZSL~\cite{romera2015embarrassingly}   & 6.6& 75.6& 12.1&    5.9& 77.8& 11.0&    12.6& 63.8& 21.0&    11.0& 27.9& 15.8 \\
		DEVISE~\cite{frome2013devise}   & 13.4& 68.7& 22.4&    17.1& 74.7& 27.8&    23.8& 53.0& 32.8&    16.9& 27.4&20.9  \\
		ALE~\cite{akata2013label}       & 16.8& 76.1& 27.5&    14.0& 81.8& 23.9&    23.7& 62.8& 34.4&    21.8& 33.1& 26.3 \\
		SAE~\cite{kodirov2017semantic}  & 1.8& 77.1& 3.5&    1.1& 82.2& 2.2&    7.8& 54.0& 13.6&    8.8& 18.0& 11.8 \\
		SYNC~\cite{changpinyo2016synthesized} & 8.9& 87.3& 16.2&    10.0& 90.5& 18.0&    11.5& 70.9& 19.8&    7.9& 43.3& 13.4 \\
	 	CVAE-ZSL~\cite{mishra2017generative}& -& -& 47.2&    -& -& 51.2&    -& -& 34.5&    -& -& 26.7  \\
	 	RelationNet~\cite{sung2018learning} & 31.4 & \textbf{91.3} & 46.7 &     30.0 &\textbf{93.4} & 45.3 &   38.1 & 61.1 & 47.0 &    - & - & - \\
	 	f-CLSWGAN~\cite{xian2018feature}   & 57.9 & 61.4& 59.6&    -& -& -&    43.7& 57.7& 49.7&    42.6& 36.6& 39.4 \\ 
	 	SGAL~\cite{yu2019zero}           & 52.7 & 75.7 & 62.2 &     55.1 & 81.2 & 65.6 &     47.1 & 44.7 & 45.9&     42.9 & 31.2 & 36.1 \\
	 	GDAN~\cite{huang2019generative}  & -&-&-&                   32.1 & 67.5 & 43.5 &     39.3 & 66.7 & 49.5&     38.1 & 89.9 &  53.4 \\
	 	CADA-VAE~\cite{schonfeld2019generalized}  &57.3 & 72.8 & 64.1 &     55.8 & 75.0 & 63.9 &     51.6 & 53.5 & 52.4&     47.2 & 35.7 & 40.6 \\
	 	f-VAEGAN-D2~\cite{xian2019f} & 57.6 & 70.6 & 63.5 &     -    & -    & -    &     48.4 & 60.1 & 53.6&     45.1 & 38.0 & 41.3 \\
	 	\hline
	 	OT-ZSL (w/o $f$)  & 57.8 & 75.2 & 65.4    &58.1 & 76.0& 65.9   & 44.2 & 56.3 & 49.5    & 44.5 &50.6 & 47.4 \\
	 	OT-ZSL     & \textbf{62.3} & 80.5& \textbf{69.6}&    \textbf{60.6}& 77.5& \textbf{68.0}&    \textbf{52.7} & \textbf{68.0}&  \textbf{59.4}&    \textbf{52.7}& \textbf{52.7}& \textbf{54.8}  \\
	\hline\hline
	\end{tabular}
	\end{small}
	\vspace{-1em}
\end{table*}

\subsection{Generalized zero-shot learning}
In the GZSL setting, seen and unseen classes are evaluated jointly when testing. 
We use the data split provided in~\cite{xian2017zero}. 
Accordingly, the testing data from the seen classes and that from the unseen classes are referred to as $\mathcal{X}_{test}^S$ and $\mathcal{X}_{test}^U$, respectively. 
Given the classifier trained over all $S+U$ classes, we evaluate the performance of different methods via the following three measurements:

1. $A_s$: average per-class top-1 accuracy on $\mathcal{X}_{test}^S$.

2. $A_u$: average per-class top-1 accuracy on $\mathcal{X}_{test}^U$.

3. $H$: harmonic mean of $A_s$ and $A_u$, $i.e.$, $H = \frac{2A_s A_u}{A_s + A_u}$.

$H$ evaluates the overall performance of the proposed method on both seen and unseen classes, which is the key criterion of GZSL problem.
The results, comparing with several state-of-the-arts, are presented in Table~\ref{res:GZSL}. 
The proposed model achieves superior performance for most of the benchmark datasets, especially on the $H$ measure. 
This demonstrates that the proposed method keeps a balance between the seen and unseen classes better than alternative approaches~\cite{norouzi2013zero, changpinyo2016synthesized}, avoiding a strong bias towards the seen classes. 
Among all methods, generative model-based ZSL approaches~\cite{mishra2017generative, verma2018generalized, xian2018feature} achieve remarkable success for the GZSL task, showing that learning a power generative model is critical for the generalization to those unseen classes. 
It can be seen that the proposed model achieves comparable, if not the best, $A_u$ over these benchmark datasets.

Note that even if we purely rely on the optimal transport distance as the objective function, and train the proposed model without the attribute-based regularizer, our method can still achieve comparable learning results to the state-of-the-art methods, as shown in the row of ``OT-ZSL (w/o $f$)'' in Table~\ref{res:GZSL}. 
This result demonstrates the effectiveness of the optimal transport framework for ZSL problems. 
On the other hand, the attribute-based regularizer is necessary to further boost the classification accuracy, which brings a non-trivial gain ($4\%\sim 10\%$) for both $A_s$ and $A_u$.

\begin{table}[t]
	\centering
    \caption{\small 
	   Top-1 accuracy in the standard ZSL problem. 
	}\label{res:SZSL}
	\begin{small}
	\begin{threeparttable}[c]
	\resizebox{\columnwidth}{!}{%
	\begin{tabular}{
	@{\hspace{3pt}}l@{\hspace{3pt}}|
	@{\hspace{3pt}}c@{\hspace{3pt}}
	@{\hspace{3pt}}c@{\hspace{3pt}}|
	@{\hspace{3pt}}c@{\hspace{3pt}}
	@{\hspace{3pt}}c@{\hspace{3pt}}|
	@{\hspace{3pt}}c@{\hspace{3pt}}
	@{\hspace{3pt}}c@{\hspace{3pt}}|
	@{\hspace{3pt}}c@{\hspace{3pt}}
	@{\hspace{3pt}}c@{\hspace{3pt}}}
	\hline\hline
	     \multirow{2}{*}{Methods}  & \multicolumn{2}{c@{\hspace{3pt}}|@{\hspace{3pt}}}{AwA1} & \multicolumn{2}{c@{\hspace{3pt}}|@{\hspace{3pt}}}{AwA2} & \multicolumn{2}{c@{\hspace{3pt}}|@{\hspace{3pt}}}{CUB}  & \multicolumn{2}{c}{SUN} \\
		&S &P
		&S &P
		&S &P 
		&S &P\\
		\hline
		SJE~\cite{akata2015evaluation}          &76.7 &65.6     &69.5 &61.9     &55.3 &53.9     &57.1 &53.7 \\
		LATEM~\cite{xian2016latent}             &74.8 &55.1     &68.7 &55.8     &49.4 &49.3     &56.9 &55.3 \\
		SSE~\cite{zhang2016learning}            &68.8 &60.1     &67.5 &61.0     &43.7 &43.9     &54.5 &51.5 \\
		ESZSL~\cite{romera2015embarrassingly}   &74.7 &58.2     &75.6 &58.6     &55.1 &53.9     &57.3 &54.5 \\
		DEVISE~\cite{frome2013devise}           &72.9 &54.2     &68.6 &59.7     &53.2 &52.0     &57.5 &56.5  \\
		ALE~\cite{akata2013label}               &78.6 &59.9     &80.3 &62.5     &53.2 &54.9     &59.1 &58.1 \\
		SAE~\cite{kodirov2017semantic}          &80.6 &53.0     &80.2 &54.1     &33.4 &33.3     &42.4 &40.3 \\
		SYNC~\cite{changpinyo2016synthesized}   &72.2 &54.0     &71.2 &46.6     &54.1 &55.6     &59.1 &56.3 \\
		CVAE-ZSL~\cite{mishra2017generative}    & - & 71.4    & -   &65.8     &-    &52.1     &-    &61.7 \\ 
		RelationNet~\cite{sung2018learning}     & - & 68.2    & -   & 64.2        & 55.6 & -    & -  & - \\
		f-CLSWGAN~\cite{xian2018feature}        & - & 68.2    & -   & -          & - & 57.3    & -  &60.8 \\
		LisGAN~\cite{li2019leveraging}          & - & 70.6    & -   & -          & -  & 58.8   & - & 61.7\\
		f-VAEGAN-D2~\cite{xian2019f}            & - & 71.1    & -   & -          & - & 61.0     &- & 65.6 \\
		LFGAA~\cite{liu2019attribute}           & - & -       & -   & 68.1       & - & 67.6     & -& 62.0 \\
		\hline
		OT-ZSL (w/o $f$)   &83.2  & 72.7     & 80.2 & 71.4        & 62.1 & 59.2  & 61.6 & 63.4 \\
	 	OT-ZSL     &\textbf{86.2} &\textbf{74.3}     &\textbf{84.5} &\textbf{74.2}     &\textbf{66.7} &\textbf{68.1}  &\textbf{66.3} &\textbf{67.9}  \\
	\hline\hline
	\end{tabular}
	}
	\begin{tablenotes}
	\begin{scriptsize}
	\item ``S'' represents the standard data split most widely used for each dataset.
	\item ``P'' is the split provided by~\cite{xian2017zero}. 
	\end{scriptsize}
	\end{tablenotes}
	\end{threeparttable}
	\end{small}
	\vspace{-1.5em}
\end{table}

\subsection{Standard zero-shot learning}
For the standard zero-shot learning setting, we synthesize samples only from the unseen classes and train the linear softmax classifier accordingly. 
In the testing phase, this classifier will be used to test the samples from the unseen classes. 
In terms of the evaluation, the average per-class top-1 accuracy is reported. Experimental results, comparing with existing ZSL approaches under two kinds of data split strategies, are listed in Table~\ref{res:SZSL}. 
We can see that the gain of our model is consistent across all the four benchmark datasets. 
Similar to GZSL, when training without the attribute-based regularizer, the performance of our method decays slightly, but still beats most of its competitors.

A large-scale experiment over ImageNet is performed as well, and the average per-class top-5 accuracy for various methods is shown in Table~\ref{res:SZSL_imagenet}. 
The improvement on ImageNet further demonstrates that the superiority of our method is consistent over scales.  

\begin{figure}[t]
    \centering
    \includegraphics[width=5.5cm,height=4cm]{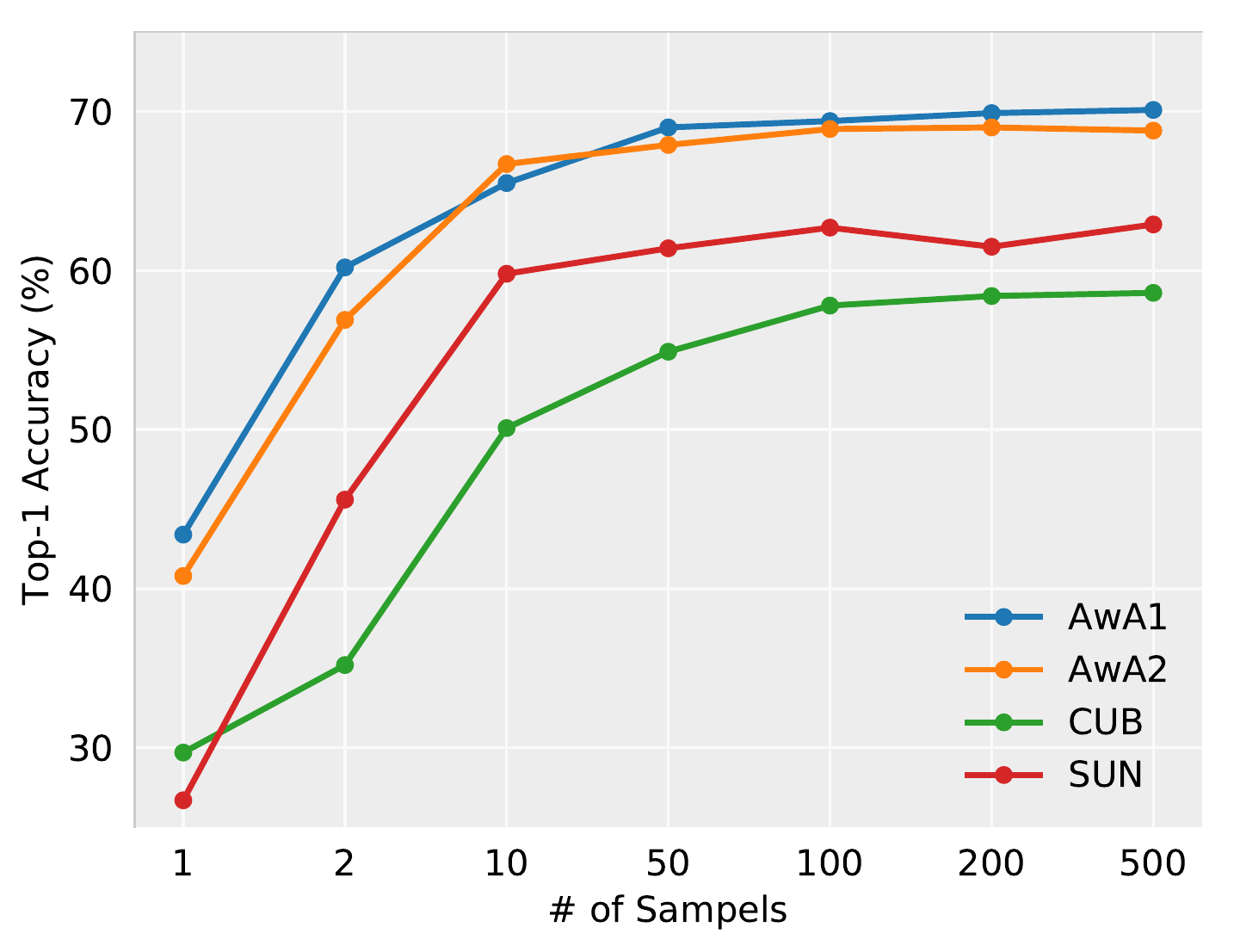}
    \caption{Top-1 Accuracy on the unseen classes with varying number of generated features in standard ZSL.}
    \label{fig:samples}
    \vspace{-1em}
\end{figure}

Additionally, we investigate the sample efficiency of our method in Figure~\ref{fig:samples}. 
For each dataset, we find that the proposed method can achieve encouraging classification accuracy even if we only generate 100 synthetic features per class to train the classifier.

\begin{table}[t]
	\centering
	\caption{\small 
		Top-5 accuracy in the standard ZSL problem of ImageNet.
	}\label{res:SZSL_imagenet}
    \begin{small}
	\begin{tabular}{l|c|c}
	\hline\hline
	    Methods &Pretrained CNN &Top-5 $A_u$\\
		\hline
		DEVISE~\cite{frome2013devise}  	&GoogleNet  & 12.8	\\
		CONSE~\cite{norouzi2013zero}    &GoogleNet  & 15.5	\\
		SS-VOC~\cite{fu2016semi}		&VGG  & 16.8	\\
		VZSL~\cite{wang2017zero}        &VGG  & 23.1	\\
		CVAE-ZSL~\cite{mishra2017generative}   &GoogleNet & 24.7	\\
		SG-GZSL~\cite{verma2018generalized} &GoogleNet & 25.4 \\ \hline
		OT-ZSL     &GoogleNet &\textbf{28.9}    \\
	\hline\hline
	\end{tabular}
    \end{small}
\end{table}

\vspace{-1em}
\subsection{Transductive zero-shot learning}
In the transductive setting, we use the standard data split for AwA1, AwA2 and CUB. 
For SUN, we follow the 707/10 data split provided by~\cite{kodirov2017semantic}. 
To yield a fair comparison relative to previous work, VGG~\cite{simonyan2014very} features are also included in this section.
The training phase of Transductive ZSL is similar to standard ZSL but with additional access to the unseen classes data, without paired label (attribute) information. 
We can directly use such unlabeled data in our optimal transport framework, $i.e.$, slightly changing Algorithm~\ref{alg2} via sampling $B_{real}$ from $\mathcal{D}_s\cup\mathcal{D}_u$ (line 4) and calculating optimal transport distance based on $B_{real}$, $B_s$ and $B_u$ (line 9).
The testing phase is the same as the standard ZSL setting. 

Table~\ref{res:TZSL} reports results for the transductive setting, with comparison to multiple state-of-the-art baselines. 
We observe that the proposed method again outperforms the other methods with a non-trivial gain --- on average, about $20\%$ improvement is achieved with the access of the unlabeled data. 
Empirically, VGG~\cite{simonyan2014very} features work slightly better than ResNet~\cite{he2016deep} feature on AwA1 and AwA2.
In the optimal-transport framework, the information of the unlabeled data is leveraged effectively, which can significantly improve the classification results. 
Again, we observe that the attribute-based regularizer helps to improve performance. 
However, because in the transductive setting we can access the unlabeled data in the training phase, the proposed optimal transport framework can find a mapping between the data and their potential attributes, even if the regularizer is not imposed.  
As a result, improvements from the regularizer in TZSL are not as significant as those in GZSL/SZSL.

\begin{table}[t]
	\centering
	\caption{\small 
	   Comparisons on $A_u$ in the transductive ZSL problem
	}\label{res:TZSL}
    \begin{small}
	\begin{tabular}{
	@{\hspace{2pt}}l@{\hspace{2pt}}|
	@{\hspace{2pt}}c@{\hspace{2pt}}|
	@{\hspace{2pt}}c@{\hspace{2pt}}
	@{\hspace{2pt}}c@{\hspace{2pt}}
	@{\hspace{2pt}}c@{\hspace{2pt}}
	@{\hspace{2pt}}c@{\hspace{2pt}}}
	\hline\hline
	     Methods & Pretrained CNN & AwA1 & AwA2 & CUB & SUN \\
		\hline
		DSRL~\cite{ye2017zero}  		&VGG &87.2 &- &57.1 &85.4  \\
		SSZSL~\cite{shojaee2016semi}    &VGG &88.6 &- &49.9 &86.2  \\
		SP-ZSR~\cite{zhang2016zero}   	&VGG &92.1 &- &55.3 &\textbf{89.5}  \\
		VZSL~\cite{wang2017zero}        &VGG &94.8 &- &66.5 &87.8  \\
		EF-ZSL~\cite{verma2017simple}   &VGG &85.2 &80.8 &60.3 & 64.5 \\
		BiDiLEL~\cite{wang2017zeroaaa}  &VGG/GoogleNet &95.0&- &62.8 &-  \\ \hline
		OT-ZSL (w/o $f$)   &VGG & 93.9 & 94.3  & 66.8 & 84.2  \\
	 	OT-ZSL     &VGG & \textbf{95.8} & \textbf{95.0} & 67.8 & 88.1   \\
		OT-ZSL (w/o $f$)   &ResNet & 93.5 & 93.6  & 67.2 & 85.8  \\
	 	OT-ZSL     &ResNet & 95.6 & 94.5 & \textbf{68.8} & 88.7   \\
	\hline\hline
	\end{tabular}
    \end{small}
    \vspace{-1em}
\end{table}

\vspace{-1em}
\subsection{IPOT vs Sinkhorn}\label{Sec: IPOTvsSinkhorn}
\begin{figure}[t]
    \centering
    \includegraphics[scale=0.37]{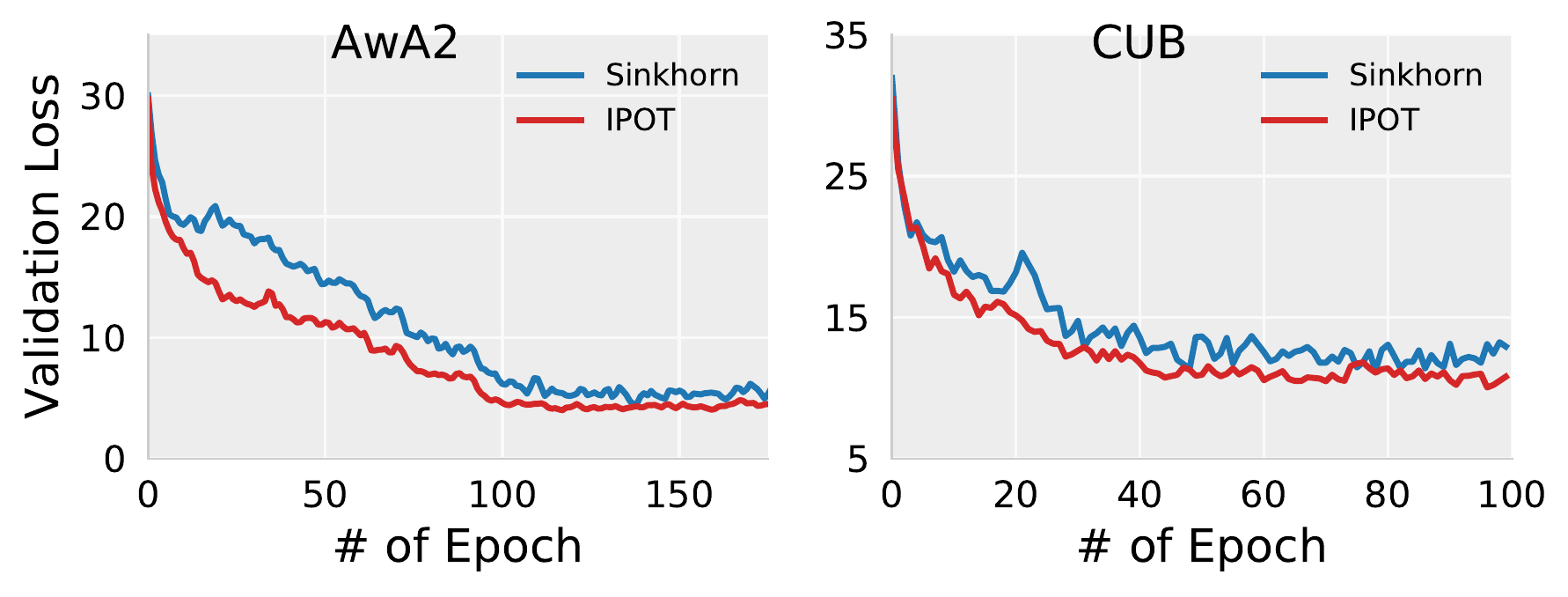}
    \caption{Measuring the validation loss w.r.t. training epoches. $\lambda=0.5$ is used for IPOT and $\lambda=0.1$, the most stable choice, for Sinkhorn iteration. }
    \label{fig:trainingPath}
    \vspace{-1em}
\end{figure}

We conduct a study on the appropriateness of using IPOT~\cite{xie2018fast} for our framework. The results are found in Figure~\ref{fig:trainingPath}. On both the AwA2 and CUB datasets, the IPOT algorithm is able to converge much faster and achieve lower losses on the validation sets. One possible reason for these observations is that IPOT is insensitive to the choice of the regularizer's weight. Therefore, IPOT is a more rational choice for our framework.

\vspace{-1em}
\subsection{Visualization}
\begin{figure}[t]
    \centering
    \includegraphics[scale=0.28]{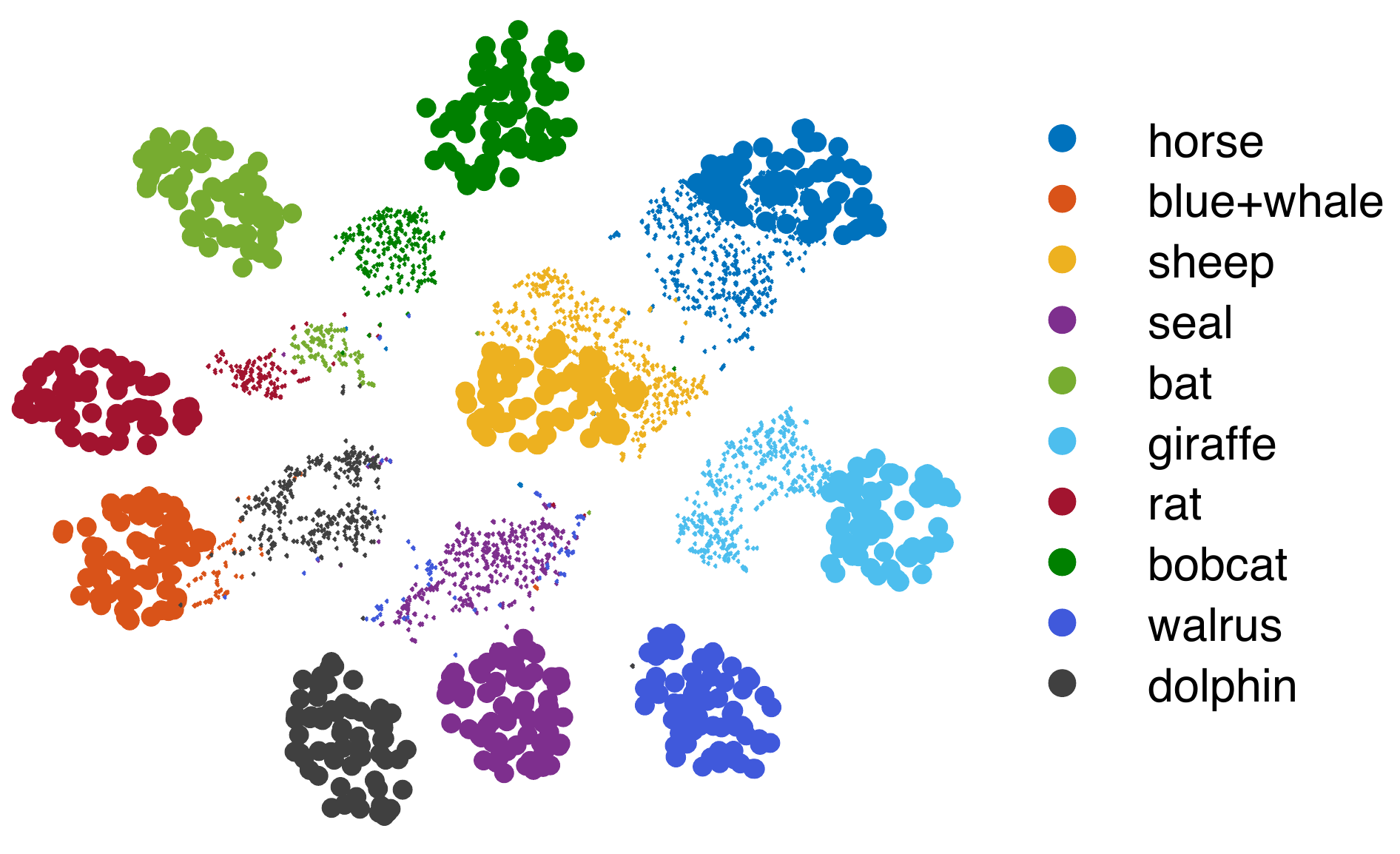}
    \caption{t-SNE visualization of the real image features and the synthetic features for the 10 unseen classes of AwA2. 
    The tiny dot points stand for the real image features while the big circles represent the synthetic features from our generative model. }
    \label{fig:tSNE}
    \vspace{-1em}
\end{figure}

To further understand our method, for the unseen classes in the AwA2 dataset we take their real features from the pretrained CNN and their synthetic features from our generator, and embed them into a 2D space using t-SNE~\cite{maaten2008visualizing}, as shown in Figure~\ref{fig:tSNE}.
The distribution of synthetic features is consistent with that of real features on the clustering structure of classes -- for each class its synthetic features are generally close to the real features within the same class. 
For example, the synthetic features on the class ``horse'' and ``sheep'' are well-mixed with the corresponding real features, demonstrating the  generalization power of our model. 
However, it should be noted that the proposed model may make mistakes in some cases, because of the natural similarity between objects, $e.g.$, the generated feature of ``blue whale'' may look more like the real feature of ``dolphin.''

\section{Conclusions}
An optimal transport framework is proposed to address zero-shot learning. 
In this framework, a conditional generator is learned to map attributes to features. 
The learning of the generator is driven by minimizing the optimal-transport distance between the distribution of generated features and that of real ones.
An attribute predictor is trained simultaneously as the generator's regularizer, encouraging the clustering structure of the generator's outputs. 
The proposed framework was developed with a focus on generalized zero-shot learning; however, we also demonstrated that it can be readily extended to standard and transductive zero-shot learning.
The proposed approach outperforms state-of-the-art methods on various datasets. 

\newpage
{\small
\bibliographystyle{ieee_fullname}
\bibliography{otzsl}

\begin{thebibliography}{10}\itemsep=-1pt

\bibitem{agueh2011barycenters}
Martial Agueh and Guillaume Carlier.
\newblock Barycenters in the {W}asserstein space.
\newblock {\em SIAM Journal on Mathematical Analysis}, 43(2):904--924, 2011.

\bibitem{akata2013label}
Zeynep Akata, Florent Perronnin, Zaid Harchaoui, and Cordelia Schmid.
\newblock Label-embedding for attribute-based classification.
\newblock In {\em CVPR}, pages 819--826. IEEE, 2013.

\bibitem{akata2015evaluation}
Zeynep Akata, Scott Reed, Daniel Walter, Honglak Lee, and Bernt Schiele.
\newblock Evaluation of output embeddings for fine-grained image
  classification.
\newblock In {\em CVPR}, pages 2927--2936, 2015.

\bibitem{altschuler2017near}
Jason Altschuler, Jonathan Weed, and Philippe Rigollet.
\newblock Near-linear time approximation algorithms for optimal transport via
  {S}inkhorn iteration.
\newblock {\em arXiv preprint arXiv:1705.09634}, 2017.

\bibitem{arjovsky2017wasserstein}
Martin Arjovsky, Soumith Chintala, and L{\'e}on Bottou.
\newblock Wasserstein generative adversarial networks.
\newblock In {\em ICML}, pages 214--223, 2017.

\bibitem{benamou2015iterative}
Jean-David Benamou, Guillaume Carlier, Marco Cuturi, Luca Nenna, and Gabriel
  Peyr{\'e}.
\newblock Iterative {B}regman projections for regularized transportation
  problems.
\newblock {\em SIAM Journal on Scientific Computing}, 37(2):A1111--A1138, 2015.

\bibitem{boissard2015distribution}
Emmanuel Boissard, Thibaut Le~Gouic, Jean-Michel Loubes, et~al.
\newblock Distribution’s template estimate with {W}asserstein metrics.
\newblock {\em Bernoulli}, 21(2):740--759, 2015.

\bibitem{changpinyo2016synthesized}
Soravit Changpinyo, Wei-Lun Chao, Boqing Gong, and Fei Sha.
\newblock Synthesized classifiers for zero-shot learning.
\newblock In {\em CVPR}, pages 5327--5336, 2016.

\bibitem{chao2016empirical}
Wei-Lun Chao, Soravit Changpinyo, Boqing Gong, and Fei Sha.
\newblock An empirical study and analysis of generalized zero-shot learning for
  object recognition in the wild.
\newblock In {\em ECCV}, pages 52--68. Springer, 2016.

\bibitem{chen2018symmetric}
Liqun Chen, Shuyang Dai, Yunchen Pu, Erjin Zhou, Chunyuan Li, Qinliang Su,
  Changyou Chen, and Lawrence Carin.
\newblock Symmetric variational autoencoder and connections to adversarial
  learning.
\newblock In {\em AISTATS}, pages 661--669, 2018.

\bibitem{chen2018zero}
Long Chen, Hanwang Zhang, Jun Xiao, Wei Liu, and Shih-Fu Chang.
\newblock Zero-shot visual recognition using semantics-preserving adversarial
  embedding networks.
\newblock In {\em CVPR}, pages 1043--1052, 2018.

\bibitem{courty2017learning}
Nicolas Courty, R{\'e}mi Flamary, and M{\'e}lanie Ducoffe.
\newblock Learning {W}asserstein embeddings.
\newblock {\em arXiv preprint arXiv:1710.07457}, 2017.

\bibitem{cuturi2013sinkhorn}
Marco Cuturi.
\newblock Sinkhorn distances: Lightspeed computation of optimal transport.
\newblock In {\em NIPS}, pages 2292--2300, 2013.

\bibitem{cuturi2014fast}
Marco Cuturi and Arnaud Doucet.
\newblock Fast computation of {W}asserstein barycenters.
\newblock In {\em ICML}, pages 685--693, 2014.

\bibitem{deng2009imagenet}
Jia Deng, Wei Dong, Richard Socher, Li-Jia Li, Kai Li, and Li Fei-Fei.
\newblock Imagenet: A large-scale hierarchical image database.
\newblock In {\em CVPR}, pages 248--255. Ieee, 2009.

\bibitem{dinu2014improving}
Georgiana Dinu, Angeliki Lazaridou, and Marco Baroni.
\newblock Improving zero-shot learning by mitigating the hubness problem.
\newblock {\em arXiv preprint arXiv:1412.6568}, 2014.

\bibitem{edwards2016towards}
Harrison Edwards and Amos Storkey.
\newblock Towards a neural statistician.
\newblock {\em arXiv preprint arXiv:1606.02185}, 2016.

\bibitem{frome2013devise}
Andrea Frome, Greg~S Corrado, Jon Shlens, Samy Bengio, Jeff Dean, Tomas
  Mikolov, et~al.
\newblock Devise: A deep visual-semantic embedding model.
\newblock In {\em NIPS}, pages 2121--2129, 2013.

\bibitem{fu2015transductive}
Yanwei Fu, Timothy~M Hospedales, Tao Xiang, and Shaogang Gong.
\newblock Transductive multi-view zero-shot learning.
\newblock {\em TPAMI}, 37(11):2332--2345, 2015.

\bibitem{fu2016semi}
Yanwei Fu and Leonid Sigal.
\newblock Semi-supervised vocabulary-informed learning.
\newblock In {\em CVPR}, pages 5337--5346, 2016.

\bibitem{genevay2017sinkhorn}
Aude Genevay, Gabriel Peyr{\'e}, and Marco Cuturi.
\newblock Sinkhorn-{A}uto{D}iff: {T}ractable {W}asserstein learning of
  generative models.
\newblock {\em arXiv preprint arXiv:1706.00292}, 2017.

\bibitem{goldberger2005neighbourhood}
Jacob Goldberger, Geoffrey~E Hinton, Sam~T Roweis, and Ruslan~R Salakhutdinov.
\newblock Neighbourhood components analysis.
\newblock In {\em NIPS}, pages 513--520, 2005.

\bibitem{goodfellow2014generative}
Ian Goodfellow, Jean Pouget-Abadie, Mehdi Mirza, Bing Xu, David Warde-Farley,
  Sherjil Ozair, Aaron Courville, and Yoshua Bengio.
\newblock Generative adversarial nets.
\newblock In {\em NIPS}, pages 2672--2680, 2014.

\bibitem{gulrajani2017improved}
Ishaan Gulrajani, Faruk Ahmed, Martin Arjovsky, Vincent Dumoulin, and Aaron~C
  Courville.
\newblock Improved training of wasserstein gans.
\newblock In {\em NIPS}, pages 5767--5777, 2017.

\bibitem{he2016deep}
Kaiming He, Xiangyu Zhang, Shaoqing Ren, and Jian Sun.
\newblock Deep residual learning for image recognition.
\newblock In {\em CVPR}, pages 770--778, 2016.

\bibitem{huang2019generative}
He Huang, Changhu Wang, Philip~S Yu, and Chang-Dong Wang.
\newblock Generative dual adversarial network for generalized zero-shot
  learning.
\newblock In {\em CVPR}, pages 801--810, 2019.

\bibitem{hubert2016learning}
Yao-Hung Hubert~Tsai, Yi-Ren Yeh, and Yu-Chiang Frank~Wang.
\newblock Learning cross-domain landmarks for heterogeneous domain adaptation.
\newblock In {\em CVPR}, pages 5081--5090, 2016.

\bibitem{kingma2014adam}
Diederik~P Kingma and Jimmy Ba.
\newblock Adam: A method for stochastic optimization.
\newblock {\em arXiv preprint arXiv:1412.6980}, 2014.

\bibitem{kingma2013auto}
Diederik~P Kingma and Max Welling.
\newblock Auto-encoding variational bayes.
\newblock {\em arXiv preprint arXiv:1312.6114}, 2013.

\bibitem{kodirov2015unsupervised}
Elyor Kodirov, Tao Xiang, Zhenyong Fu, and Shaogang Gong.
\newblock Unsupervised domain adaptation for zero-shot learning.
\newblock In {\em ICCV}, pages 2452--2460, 2015.

\bibitem{kodirov2017semantic}
Elyor Kodirov, Tao Xiang, and Shaogang Gong.
\newblock Semantic autoencoder for zero-shot learning.
\newblock {\em arXiv preprint arXiv:1704.08345}, 2017.

\bibitem{lampert2014attribute}
Christoph~H Lampert, Hannes Nickisch, and Stefan Harmeling.
\newblock Attribute-based classification for zero-shot visual object
  categorization.
\newblock {\em TPAMI}, 36(3):453--465, 2014.

\bibitem{lee2018multi}
Chung-Wei Lee, Wei Fang, Chih-Kuan Yeh, and Yu-Chiang Frank~Wang.
\newblock Multi-label zero-shot learning with structured knowledge graphs.
\newblock In {\em CVPR}, pages 1576--1585, 2018.

\bibitem{li2019leveraging}
Jingjing Li, Mengmeng Jing, Ke Lu, Zhengming Ding, Lei Zhu, and Zi Huang.
\newblock Leveraging the invariant side of generative zero-shot learning.
\newblock In {\em CVPR}, pages 7402--7411, 2019.

\bibitem{li2015semi}
Xin Li, Yuhong Guo, and Dale Schuurmans.
\newblock Semi-supervised zero-shot classification with label representation
  learning.
\newblock In {\em CVPR}, pages 4211--4219, 2015.

\bibitem{li2018discriminative}
Yan Li, Junge Zhang, Jianguo Zhang, and Kaiqi Huang.
\newblock Discriminative learning of latent features for zero-shot recognition.
\newblock In {\em CVPR}, pages 7463--7471, 2018.

\bibitem{liu2019attribute}
Yang Liu, Jishun Guo, Deng Cai, and Xiaofei He.
\newblock Attribute attention for semantic disambiguation in zero-shot
  learning.
\newblock In {\em ICCV}, pages 6698--6707, 2019.

\bibitem{long2015learning}
Mingsheng Long, Yue Cao, Jianmin Wang, and Michael~I Jordan.
\newblock Learning transferable features with deep adaptation networks.
\newblock {\em arXiv preprint arXiv:1502.02791}, 2015.

\bibitem{maaten2008visualizing}
Laurens van~der Maaten and Geoffrey Hinton.
\newblock Visualizing data using t-sne.
\newblock {\em JMLR}, 9(Nov):2579--2605, 2008.

\bibitem{mikolov2013distributed}
Tomas Mikolov, Ilya Sutskever, Kai Chen, Greg~S Corrado, and Jeff Dean.
\newblock Distributed representations of words and phrases and their
  compositionality.
\newblock In {\em NIPS}, pages 3111--3119, 2013.

\bibitem{mishra2017generative}
Ashish Mishra, M Reddy, Anurag Mittal, and Hema~A Murthy.
\newblock A generative model for zero shot learning using conditional
  variational autoencoders.
\newblock {\em arXiv preprint arXiv:1709.00663}, 2017.

\bibitem{norouzi2013zero}
Mohammad Norouzi, Tomas Mikolov, Samy Bengio, Yoram Singer, Jonathon Shlens,
  Andrea Frome, Greg~S Corrado, and Jeffrey Dean.
\newblock Zero-shot learning by convex combination of semantic embeddings.
\newblock {\em arXiv preprint arXiv:1312.5650}, 2013.

\bibitem{pele2009fast}
Ofir Pele and Michael Werman.
\newblock Fast and robust earth mover's distances.
\newblock In {\em CVPR}, pages 460--467. IEEE, 2009.

\bibitem{peyre2017computational}
Gabriel Peyr{\'e}, Marco Cuturi, et~al.
\newblock Computational optimal transport.
\newblock Technical report, 2017.

\bibitem{ravi2016optimization}
Sachin Ravi and Hugo Larochelle.
\newblock Optimization as a model for few-shot learning.
\newblock 2016.

\bibitem{rohrbach2013transfer}
Marcus Rohrbach, Sandra Ebert, and Bernt Schiele.
\newblock Transfer learning in a transductive setting.
\newblock In {\em NIPS}, pages 46--54, 2013.

\bibitem{romera2015embarrassingly}
Bernardino Romera-Paredes and Philip Torr.
\newblock An embarrassingly simple approach to zero-shot learning.
\newblock In {\em ICML}, pages 2152--2161, 2015.

\bibitem{russakovsky2015imagenet}
Olga Russakovsky, Jia Deng, Hao Su, Jonathan Krause, Sanjeev Satheesh, Sean Ma,
  Zhiheng Huang, Andrej Karpathy, Aditya Khosla, Michael Bernstein, et~al.
\newblock Imagenet large scale visual recognition challenge.
\newblock {\em IJCV}, 115(3):211--252, 2015.

\bibitem{salimans2018improving}
Tim Salimans, Han Zhang, Alec Radford, and Dimitris Metaxas.
\newblock Improving gans using optimal transport.
\newblock {\em arXiv preprint arXiv:1803.05573}, 2018.

\bibitem{schmitz2017wasserstein}
Morgan~A Schmitz, Matthieu Heitz, Nicolas Bonneel, Fred Ngole, David
  Coeurjolly, Marco Cuturi, Gabriel Peyr{\'e}, and Jean-Luc Starck.
\newblock Wasserstein dictionary learning: {O}ptimal transport-based
  unsupervised nonlinear dictionary learning.
\newblock {\em SIAM Journal on Imaging Sciences}, 11(1):643--678, 2018.

\bibitem{schonfeld2019generalized}
Edgar Schonfeld, Sayna Ebrahimi, Samarth Sinha, Trevor Darrell, and Zeynep
  Akata.
\newblock Generalized zero-and few-shot learning via aligned variational
  autoencoders.
\newblock In {\em CVPR}, pages 8247--8255, 2019.

\bibitem{shojaee2016semi}
Seyed~Mohsen Shojaee and Mahdieh~Soleymani Baghshah.
\newblock Semi-supervised zero-shot learning by a clustering-based approach.
\newblock {\em arXiv preprint arXiv:1605.09016}, 2016.

\bibitem{simonyan2014very}
Karen Simonyan and Andrew Zisserman.
\newblock Very deep convolutional networks for large-scale image recognition.
\newblock {\em arXiv preprint arXiv:1409.1556}, 2014.

\bibitem{sinkhorn1967concerning}
Richard Sinkhorn and Paul Knopp.
\newblock Concerning nonnegative matrices and doubly stochastic matrices.
\newblock {\em Pacific Journal of Mathematics}, 21(2):343--348, 1967.

\bibitem{socher2013zero}
Richard Socher, Milind Ganjoo, Christopher~D Manning, and Andrew Ng.
\newblock Zero-shot learning through cross-modal transfer.
\newblock In {\em NIPS}, pages 935--943, 2013.

\bibitem{sung2018learning}
Flood Sung, Yongxin Yang, Li Zhang, Tao Xiang, Philip~HS Torr, and Timothy~M
  Hospedales.
\newblock Learning to compare: Relation network for few-shot learning.
\newblock In {\em Proceedings of the IEEE Conference on Computer Vision and
  Pattern Recognition}, pages 1199--1208, 2018.

\bibitem{szegedy2015going}
Christian Szegedy, Wei Liu, Yangqing Jia, Pierre Sermanet, Scott Reed, Dragomir
  Anguelov, Dumitru Erhan, Vincent Vanhoucke, and Andrew Rabinovich.
\newblock Going deeper with convolutions.
\newblock In {\em CVPR}, pages 1--9, 2015.

\bibitem{tsai2017learning}
Yao-Hung~Hubert Tsai, Liang-Kang Huang, and Ruslan Salakhutdinov.
\newblock Learning robust visual-semantic embeddings.
\newblock {\em arXiv preprint arXiv:1703.05908}, 2017.

\bibitem{verma2018generalized}
V~Kumar Verma, Gundeep Arora, Ashish Mishra, and Piyush Rai.
\newblock Generalized zero-shot learning via synthesized examples.
\newblock In {\em CVPR}, 2018.

\bibitem{verma2017simple}
Vinay~Kumar Verma and Piyush Rai.
\newblock A simple exponential family framework for zero-shot learning.
\newblock In {\em Joint European Conference on Machine Learning and Knowledge
  Discovery in Databases}, pages 792--808. Springer, 2017.

\bibitem{villani2008optimal}
C{\'e}dric Villani.
\newblock {\em Optimal transport: old and new}, volume 338.
\newblock Springer Science \& Business Media, 2008.

\bibitem{vincent2010stacked}
Pascal Vincent, Hugo Larochelle, Isabelle Lajoie, Yoshua Bengio, and
  Pierre-Antoine Manzagol.
\newblock Stacked denoising autoencoders: Learning useful representations in a
  deep network with a local denoising criterion.
\newblock {\em JMLR}, 11(Dec):3371--3408, 2010.

\bibitem{vinyals2016matching}
Oriol Vinyals, Charles Blundell, Tim Lillicrap, Daan Wierstra, et~al.
\newblock Matching networks for one shot learning.
\newblock In {\em NIPS}, pages 3630--3638, 2016.

\bibitem{wah2011caltech}
Catherine Wah, Steve Branson, Peter Welinder, Pietro Perona, and Serge
  Belongie.
\newblock The caltech-ucsd birds-200-2011 dataset.
\newblock 2011.

\bibitem{wang2017zeroaaa}
Qian Wang and Ke Chen.
\newblock Zero-shot visual recognition via bidirectional latent embedding.
\newblock {\em IJCV}, 124(3):356--383, 2017.

\bibitem{wang2017zero}
Wenlin Wang, Yunchen Pu, Vinay~Kumar Verma, Kai Fan, Yizhe Zhang, Changyou
  Chen, Piyush Rai, and Lawrence Carin.
\newblock Zero-shot learning via class-conditioned deep generative models.
\newblock {\em AAAI}, 2018.

\bibitem{wang2018zero}
Xiaolong Wang, Yufei Ye, and Abhinav Gupta.
\newblock Zero-shot recognition via semantic embeddings and knowledge graphs.
\newblock In {\em CVPR}, pages 6857--6866, 2018.

\bibitem{xian2016latent}
Yongqin Xian, Zeynep Akata, Gaurav Sharma, Quynh Nguyen, Matthias Hein, and
  Bernt Schiele.
\newblock Latent embeddings for zero-shot classification.
\newblock In {\em CVPR}, pages 69--77, 2016.

\bibitem{xian2018feature}
Yongqin Xian, Tobias Lorenz, Bernt Schiele, and Zeynep Akata.
\newblock Feature generating networks for zero-shot learning.
\newblock In {\em CVPR}, 2018.

\bibitem{xian2017zero}
Yongqin Xian, Bernt Schiele, and Zeynep Akata.
\newblock Zero-shot learning-the good, the bad and the ugly.
\newblock {\em arXiv preprint arXiv:1703.04394}, 2017.

\bibitem{xian2019f}
Yongqin Xian, Saurabh Sharma, Bernt Schiele, and Zeynep Akata.
\newblock f-vaegan-d2: A feature generating framework for any-shot learning.
\newblock In {\em Proceedings of the IEEE Conference on Computer Vision and
  Pattern Recognition}, pages 10275--10284, 2019.

\bibitem{xiao2010sun}
Jianxiong Xiao, James Hays, Krista~A Ehinger, Aude Oliva, and Antonio Torralba.
\newblock Sun database: Large-scale scene recognition from abbey to zoo.
\newblock In {\em CVPR}, pages 3485--3492. IEEE, 2010.

\bibitem{xie2018fast}
Yujia Xie, Xiangfeng Wang, Ruijia Wang, and Hongyuan Zha.
\newblock A fast proximal point method for wasserstein distance.
\newblock {\em arXiv preprint arXiv:1802.04307}, 2018.

\bibitem{ye2017fast}
Jianbo Ye, Panruo Wu, James~Z Wang, and Jia Li.
\newblock Fast discrete distribution clustering using {W}asserstein barycenter
  with sparse support.
\newblock {\em TSP}, 65(9):2317--2332, 2017.

\bibitem{ye2017zero}
Meng Ye and Yuhong Guo.
\newblock Zero-shot classification with discriminative semantic representation
  learning.
\newblock In {\em CVPR}, 2017.

\bibitem{yu2019zero}
Hyeonwoo Yu and Beomhee Lee.
\newblock Zero-shot learning via simultaneous generating and learning.
\newblock In {\em NIPS}, pages 46--56, 2019.

\bibitem{zemel2017fr}
Yoav Zemel and Victor~M Panaretos.
\newblock Fr{\'e}chet means and {P}rocrustes analysis in {W}asserstein space.
\newblock {\em arXiv preprint arXiv:1701.06876}, 2017.

\bibitem{zhang2015zero}
Ziming Zhang and Venkatesh Saligrama.
\newblock Zero-shot learning via semantic similarity embedding.
\newblock In {\em ICCV}, 2015.

\bibitem{zhang2016learning}
Ziming Zhang and Venkatesh Saligrama.
\newblock Learning joint feature adaptation for zero-shot recognition.
\newblock {\em arXiv preprint arXiv:1611.07593}, 2016.

\bibitem{zhang2016zero}
Ziming Zhang and Venkatesh Saligrama.
\newblock Zero-shot learning via joint latent similarity embedding.
\newblock In {\em CVPR}, pages 6034--6042, 2016.

\end{thebibliography}
}

\clearpage
\appendix
\section{IPOT}

A detailed description of the IPOT algorithm used in our framework is summarized in Algorithm~\ref{alg1}, where ``$\odot$'' is the Hadamard product and ``$\frac{\cdot}{\cdot}$'' represents element-wise division. 
It is notable that this method works well with batch-based optimization, $i.e.$, $N$ in Algorithm~\ref{alg1} can represent either the size of the whole dataset or the size of a batch.

Specifically, compared with the Sinkhorn iteration algorithm, IPOT {\em changes} the objective function by adding an entropy regularizer on $\Tv$.
Although such a modification converts the optimal transport problem to be strictly convex, its success is highly dependent on the choice of regularizer weight. On one hand, if the weight is too large, Sinkhorn iteration only obtains an over-smoothed $\Tv$ with a large number of iterations. 
On the other hand, if the weight is too small, Sinkhorn iteration suffers from numerical-stability issues. 
IPOT, in contrast, solves the original optimal transport problem.
In particular, the regularizer in (\ref{eq:ipot}) just controls the learning process and its weight $\lambda$ mainly affects the convergence rate. 
If we reduce its weight $\lambda$ with respect to the number of iterations, the final result $\Tv^*$ will be equal to that obtained by solving (\ref{eq:ot2}) directly. 
Additionally, because the weight $\lambda$ mainly affects convergence rate, we can choose it in a wide range to achieve better numerical stability than Sinkhorn iteration.

\begin{algorithm}[h]
  \caption{IPOT Algorithm}
  \label{alg1}
\begin{algorithmic}[1]
    \STATE \textbf{Input:} Real features $\{\xv_n \}_{n=1}^N$, generated features $\{\hat{\xv}_m\}_{m=1}^M$, $\lambda=0.5$, $\muv=[\frac{1}{N}]$, $\vv=[\frac{1}{M}]$.
    \STATE \textbf{Output:} Optimal transport $\Tv^*$
    \STATE Calculate $\Cv=[C_{nm}]$, with $C_{nm} = 1 - \frac{\xv_n^{\top} \hat{\xv}_m}{\Vert\xv_n\Vert_2 \Vert\hat{\xv}_m\Vert_2}$.
    \STATE $\Gv = \exp(-\frac{\Cv}{\lambda})$.
    \STATE Initialize $\av=\muv$, $\Tv^{(1)} = \muv\vv^{\top}$
    \FOR{$t=1,...,T$}
    \STATE $\Kv = \Gv\odot \Tv^{(t)}$ 
    \STATE \CommentSty{Sinkhorn-Knopp Algorithm:}
    \FOR{$j = 1, ..., J$} 
    \STATE $\bv=\frac{\vv}{\Kv^{\top}\av}$ and $\av=\frac{\muv}{\Kv\bv}$.
    \ENDFOR
    \STATE $\Tv^{(t+1)}=\text{diag}(\av) \Kv\text{diag}(\bv)$
    \ENDFOR
    \STATE $\Tv^*=\Tv^{(T+1)}$
\end{algorithmic}
\end{algorithm}

\end{document}